\documentclass{article} \usepackage{iclr2022_conference,times}
\pdfoutput=1

\usepackage{amsmath,amsfonts,bm}

\def\eqref#1{equation~\ref{#1}}

\def\1{\bm{1}}

\def\vtheta{{\bm{\theta}}}

\DeclareMathAlphabet{\mathsfit}{\encodingdefault}{\sfdefault}{m}{sl}
\SetMathAlphabet{\mathsfit}{bold}{\encodingdefault}{\sfdefault}{bx}{n}

\usepackage{listings}
\usepackage{hyperref}
\usepackage[noabbrev,capitalize,nameinlink]{cleveref}
\usepackage{url}
\usepackage{booktabs}
\usepackage{graphicx}
\usepackage{algorithm}
\usepackage[noend]{algpseudocode}
\usepackage{multirow}

\crefname{table}{Table}{}
\crefname{figure}{Figure}{}
\crefname{algorithm}{Algorithm}{}
\crefname{equation}{equation}{}
\crefname{appendix}{Appendix}{}

\newcommand{\ptheta}{p_{\vtheta}}

\newcommand{\xx}{\mathbf{x}}
\newcommand{\yy}{\mathbf{y}}

\newcommand{\monotext}{{\cal M}}
\newcommand{\bitextback}{{\cal B}_\textit{back}}

\iclrfinalcopy
\title{Unsupervised neural machine translation \\ with generative language models only}

\author{\hspace{52mm}Jesse Michael Han \AND Igor Babuschkin \And Harrison Edwards
  \And Arvind Neelakantan \And Tao Xu \And Stanislas Polu \And Alex
  Ray \And Pranav Shyam \And Aditya Ramesh \And Alec Radford \And Ilya
  Sutskever \AND \\ \hspace{60mm}{\large OpenAI}
}

\graphicspath{{images}}
\begin{document}

\maketitle

\begin{abstract}
      We show how to derive
  state-of-the-art unsupervised neural machine translation systems
  from generatively pre-trained language models. Our method
  consists of three steps: \emph{few-shot amplification}, \emph{distillation}, and
  \emph{backtranslation}.  We first use the zero-shot translation
  ability of large pre-trained language
  models to generate translations for a small set of unlabeled
  sentences.  We then amplify these zero-shot translations by using them as
  few-shot demonstrations for sampling a larger synthetic dataset.
          This dataset is distilled by
  discarding the few-shot demonstrations and then fine-tuning. During
  backtranslation, we repeatedly generate translations for a set of
  inputs and then fine-tune a single
  language model on both directions of the translation task at once,
  ensuring cycle-consistency by swapping the roles of gold
  monotext and generated translations when fine-tuning.     By using
  our method to leverage GPT-3's zero-shot
  translation capability, we achieve a new state-of-the-art in
  unsupervised translation on the WMT14 English-French benchmark,
  attaining a BLEU score of 42.1.
\end{abstract}

\section{Introduction}
Recent work on generative pre-training has shown
that with sufficient data and
scale
\citep{DBLP:journals/corr/abs-2001-08361,DBLP:journals/corr/abs-2010-14701},
large language models (LMs) can learn a diverse suite of tasks without
explicit supervision \citep{radford2019language}, and that even
stronger performance on these tasks can be elicited using few-shot
demonstrations \citep{DBLP:conf/nips/BrownMRSKDNSSAA20}. While
few-shot prompting is flexible and
enables strong performance on a diverse suite of NLP tasks
to be coaxed out of generatively pre-trained LMs without further
fine-tuning, its benefits are
most pronounced with larger models, with commensurate training,
inference, compute, and data costs. Furthermore, the very generality of the
pre-training objective which enables multi-task learning can produce LMs
with more knowledge than is immediately apparent,
requiring carefully designed prompts to bring out fully. The desire to
unlock and amplify these latent abilities while also reducing the cost
of few-shot prompting motivates our present work,
which allows us to continue
fine-tuning our models, obtaining more performance from
smaller models and pushing our larger models even further, without
resorting to few-shot prompting at test time or any additional supervision
at train~time.

We target the domain of unsupervised
  neural machine translation (NMT), which typically involves
\emph{bootstrapping} a weak translation model before amplifying its translation ability via
\emph{backtranslation}. Recent work in unsupervised NMT has been
dominated by large encoder-decoder architectures where the bootstrap is
implemented by denoising/autoencoding tasks
(\emph{e.g.}, multilingual Cloze \citep{DBLP:conf/naacl/DevlinCLT19, DBLP:conf/nips/ConneauL19}, masked-span
prediction \citep{DBLP:journals/jmlr/RaffelSRLNMZLL20, DBLP:conf/naacl/XueCRKASBR21},
reconstruction from corrupted inputs
\citep{DBLP:conf/emnlp/WangZJLL19, DBLP:journals/tacl/LiuGGLEGLZ20})
intended to produce strong encoders and
aligned multilingual representations for decoding.
In our present work, we show that generative language modeling alone
can implement the entire unsupervised NMT pipeline, and derive
state-of-the-art unsupervised NMT systems using only generatively
pre-trained language models.
We implement the bootstrap by first sampling a small number of
zero-shot translations from GPT-3. These are then used as
few-shot prompts to sample a larger dataset of synthetic translations.
The few-shot prompts are then discarded and the generated
samples are \emph{distilled} by fine-tuning the model on
these synthetic data in the zero-shot format. This produces a language model
aligned to our translation format and amenable to large-scale
backtranslation. By using our method to leverage GPT-3's zero-shot
translation capability, we achieve a new state-of-the-art in
unsupervised translation on the WMT14 English-French benchmark,
attaining a BLEU score of 42.1.

\section{Background and related work} \label{sec:background}
  The modern approach to unsupervised neural machine translation typically involves
encoder-decoder architectures jointly trained via
denoising autoencoding / reconstruction tasks \citep{DBLP:conf/icml/VincentLBM08,
  DBLP:conf/nips/ConneauL19,DBLP:journals/tacl/LiuGGLEGLZ20,DBLP:journals/corr/abs-2012-15547,
DBLP:journals/jmlr/RaffelSRLNMZLL20,DBLP:conf/naacl/XueCRKASBR21,
DBLP:conf/emnlp/WangZJLL19, DBLP:journals/tacl/LiuGGLEGLZ20, DBLP:conf/icml/SongTQLL19} and
backtranslation \citep{DBLP:conf/acl/SennrichHB16,
  DBLP:conf/emnlp/EdunovOAG18,
  DBLP:journals/corr/abs-1806-04402}. This approach to unsupervised
NMT is codified by
\citet{DBLP:conf/iclr/ArtetxeLAC18} and
\citet{DBLP:conf/iclr/LampleCDR18},
although various ideas can be
traced back further: unsupervised machine translation was framed as a
deciphering task by \citet{DBLP:conf/acl/RaviK11} and backtranslation
was first introduced for machine translation as a method for
data augmentation using target-side monolingual data by
\citet{DBLP:conf/acl/SennrichHB16}. Denoising autoencoding with a bilingual encoder can be
viewed as a kind of latent bilingual lexicon induction, necessary for
producing sufficiently aligned embeddings to
kick-start backtranslation; such techniques have been extensively studied
in the context of machine translation
\citep{DBLP:conf/acl/ArtetxeLA17, DBLP:conf/coling/KlementievTB12,
DBLP:conf/acl/VulicM15, DBLP:journals/corr/HuYLSX17, DBLP:conf/nips/GoyalLZZCB16, DBLP:conf/nips/ShenLBJ17}.

At the same time, recent work on large-scale generative
pre-training has
demonstrated that with sufficient data and model
scale \citep{DBLP:journals/corr/abs-2001-08361,DBLP:journals/corr/abs-2010-14701}, transformer language models begin learning a variety of tasks
without explicit supervision \citep{radford2019language} and that even
stronger performance can be
coaxed from them using few-shot prompts \citep{DBLP:conf/nips/BrownMRSKDNSSAA20}. Our
present work unifies these lines of research by using generative
language modeling to
simplify unsupervised NMT even
further: we show how with sufficient scale, pre-training, and
clever prompting, a single generative language model can implement
the entire unsupervised neural machine translation pipeline, avoiding
optimizations such as denoising
autoencoding, auxiliary / adversarial losses in latent space, or
ad-hoc bilingual dictionaries.

Our reliance on large-scale generative pre-training is similar to
prior work in unsupervised NMT which uses large-scale language
modeling tasks on internet data as part of the bootstrap
\citep{DBLP:conf/nips/ConneauL19, DBLP:conf/acl/ConneauKGCWGGOZ20,
  DBLP:journals/tacl/LiuGGLEGLZ20}.
The role of few-shot prompting and distillation in our method is related to recent
work on unsupervised data augmentation using language models
\citep{DBLP:conf/aaai/Anaby-TavorCGKK20,
  DBLP:journals/corr/abs-2103-00453,
  DBLP:journals/corr/abs-2003-02245,
  DBLP:journals/corr/abs-2004-13845,
  DBLP:journals/corr/abs-2104-07540, DBLP:conf/emnlp/YangMFSBWBCD20}
and is also in the same spirit as recent work on \emph{self-training}
and \emph{noisy-student training}
\citep{DBLP:journals/corr/abs-2108-12589,
  DBLP:journals/corr/abs-2109-06270,
  DBLP:conf/cvpr/XieLHL20}. Recent work on scaling laws for neural
machine translation has shown that transformer decoders exhibit more
favorable scaling than encoders \citep{DBLP:journals/corr/abs-2109-07740}. The
few-shot distillation component of our
method bears some resemblance to contemporaneous work by
\citet{DBLP:journals/corr/abs-2109-09193} which uses few-shot
prompting for unsupervised data
augmentation, though they focus only on
inference for text classification rather than generation for
sequence-to-sequence tasks like machine translation and do not study
the phenomena of self-amplification nor
few-shot data efficiency (\Cref{sec:discussion}) as we do.

\section{Backtranslation via language modeling} \label{sec:bts}

\begin{algorithm}[!h]
  \begin{algorithmic}[1]
  	\Require{Source monotext $\monotext_S$; target monotext
          $\monotext_T$; number of iterations \(I\); number of samples
          per
        iteration \(J\); monotext formatter
        \(f(\cdot)\); bitext formatter
        \(g(\cdot, \cdot)\); parameters \(\vtheta\) of language model
        \(p_{\vtheta}(\cdot)\) trained to complete outputs of \(f\) to
        outputs of \(g\).}
      \vspace{1mm}
    \Ensure{Final model parameters $\vtheta$.}
    \vspace{1mm}
    \For{$i=1$ \textbf{to} $I$}
    \State $\bitextback \gets \emptyset$
    \For{$j=1$ \textbf{to} $J$}
        \State \(\yy \sim \monotext_S \cup \monotext_T\)
    	\State $\tilde{\xx} \sim \ptheta(\cdot \mid f(\yy))$
    	\State $\bitextback \gets \bitextback \cup \{\langle \tilde{\xx},
        \yy \rangle \}$
    \EndFor
    \State {estimate} $\vtheta$ {by} {maximizing} $\log \ptheta$ {of}
    $g(\tilde{\xx}, \yy)$ for \(\langle \tilde{\xx}, \yy \rangle \in
    \bitextback\)
    \EndFor

\end{algorithmic}
  \caption{{Iterated backtranslation using a single generative
language model}}
  \label{alg:backtranslation}
\end{algorithm}

Backtranslation was first introduced in the context of machine
translation as a method for data augmentation
using target-side monolingual data by
sampling synthetic source-to-target data from another target-to-source
translation model \citep{DBLP:conf/wmt/BojarT11,
  DBLP:conf/acl/SennrichHB16, DBLP:journals/corr/abs-1804-06189}. In
our present work, we cast machine translation as
a language modeling task and jointly train and sample generations from
a single language model for both source-to-target and target-to-source
translation.

\label{sec:org5115ca6}
Given bitext \texttt{⟨seq1, seq2⟩} in languages \(L_1\) and \(L_2\), we format the translation task as follows:
\[
  \texttt{[L1] <seq1> [[TRANSLATE]] [L2] <seq2>}
\]

At test-time, the LM is prompted with \texttt{[L1] <seq>
  [[TRANSLATE]] [L2]} and we parse a candidate translation
\texttt{<sampledSeq>} from the sampled completion. Backtranslation is
implemented by reversing the roles of \texttt{seq} and
\texttt{sampledSeq} and fine-tuning on the bitext
\texttt{⟨sampledSeq, seq⟩}.

We remark that in contrast to the
interpretation of backtranslation as a wake-sleep
algorithm \citep{DBLP:journals/corr/abs-1806-04402}, where the forwards
and backwards translators are trained alternately, we use a single
language model for both forwards and backwards translation and train
on both directions jointly at every iteration. 
There are various ways to train a model using
backtranslation, \emph{e.g.}, completely online (interleaving minibatch gradient updates
and sampling) versus offline (backtranslating the entire training dataset
at each epoch; potentially re-training the model from scratch after
sampling new backtranslations). In practice, we find that data scaling
of a model's optimal test loss and BLEU score quickly saturates
on backtranslations from previous versions of the model, and opt for
a semi-online setup where we synchronously sample a relatively small
number of \(L_1\)-\(L_2\) and \(L_2\)-\(L_1\) pairs before resuming
training for a single epoch on the newly sampled data. We refer to
this as a single \emph{iteration} of backtranslation.

Formally, \Cref{alg:backtranslation} describes our implemention of
backtranslation using a single generative language model
\(p_{\vtheta}(\cdot)\). We assume that \(p_{\vtheta}(\cdot)\) has
already been trained to complete formatted monotext (\texttt{[L1] <seq1> [[TRANSLATE]] [L2]}) to formatted
bitext (\texttt{[L1] <seq1> [[TRANSLATE]] [L2] <seq2>}).

\section{The bootstrap: generative pre-training, few-shot amplification, and
  distillation} \label{sec:bootstrap}

\begin{figure*}[!htbp] \label{fig:bootstrap-diagram}
  \begin{center}
    \includegraphics[width=0.95\textwidth]{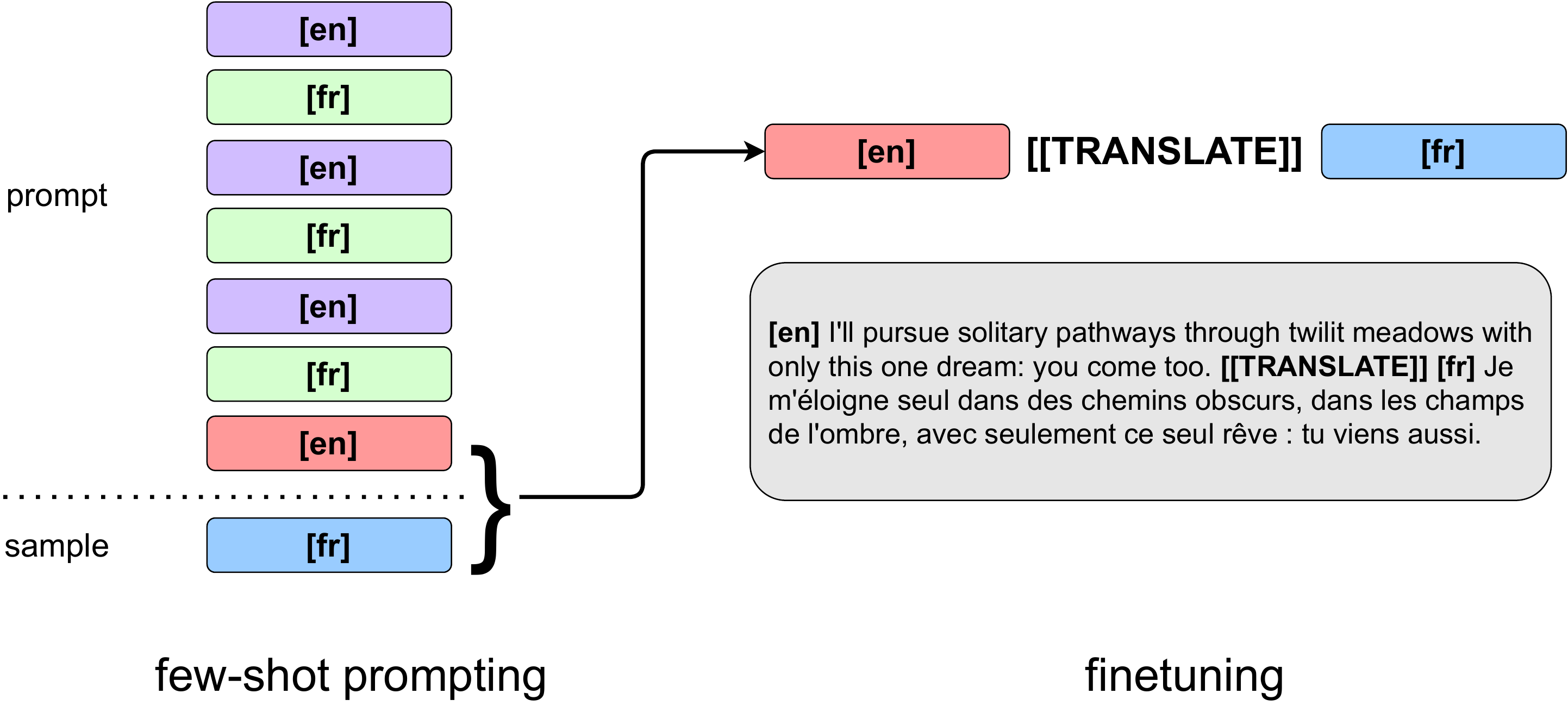}
  \end{center}
  \caption{Illustration of our bootstrap procedure, which we call
    \emph{few-shot distillation}. We use few-shot prompts sampled from
    GPT-3 to generate an
  initial dataset of synthetic translations from a
  generatively pre-trained language model (left). The few-shot examples
are then discarded and the synthetic bitext reformatted for fine-tuning
on the autoregressive language modeling objective (right).}
  \end{figure*}

The modern approach to unsupervised NMT is parametrized by a choice of
initialization or \emph{bootstrap}. The bootstrap has typically relied on
some form of unsupervised cross-lingual representation learning,
\emph{e.g.}, bilingual dictionaries initialized from unsupervised
cross-lingual word embeddings \citep{DBLP:conf/iclr/LampleCDR18,
  DBLP:conf/iclr/ArtetxeLAC18} or multilingual masked language
modeling followed by denoising autoencoding with a shared
encoder and decoder \citep{DBLP:conf/nips/ConneauL19}.

In \Cref{sec:bts}, we formulated iterative backtranslation in terms of
language modeling, assuming a language model which has already been
trained to follow a particular instruction format for translation. To
complete our procedure, we must supply such a language model. Unlike previous
work on unsupervised NMT, we use language models from the GPT-3 family
\citep{DBLP:conf/nips/BrownMRSKDNSSAA20} which have
been \emph{generatively pre-trained} on a large corpus of Internet
data. A key observation from the body of work around GPT-3 is that
generative pre-training at scale induces strong in-context
metalearning abilities, two special cases of which are (1) instruction
following and (2) few-shot
prompting:
a sufficiently trained large language model benefits from both detailed
natural language descriptions of tasks and, when given in-context
examples, can achieve strong performance on a diverse suite of
tasks (\emph{e.g.}, question-answering, natural language inference,
translation.) We implement the bootstrap by exploiting both of these
abilities, by using natural language instruction to produce zero-shot
translations and few-shot prompting during amplification.

\subsection{Few-shot amplification and distillation} It thus remains to \label{para:bootstrap}
adapt our generatively pre-trained models' few-shot translation
ability to the zero-shot format specified in \Cref{sec:bts}. We do
this in a two-stage process. We first sample a small number of
zero-shot translations from GPT-3. Given bitext \texttt{⟨srcSeq, tgtSeq⟩} in
\texttt{srcLang} and \texttt{tgtLang}, and a
stop-sequence \texttt{<sep>}, we use the following format for
zero-shot prompting:
\begin{lstlisting}
  <sep> Given the following passage in <srcLang>: <sep> <srcSeq> <sep>
  a good <tgtLang> translation is: <sep> <tgtSeq> <sep>.
\end{lstlisting}
At test-time, we sample a
completion until the stop-sequence \texttt{<sep>} is detected;
throughout we set \texttt{<sep>} to be
\texttt{{\textbackslash}n---{\textbackslash}n}.

We \emph{amplify} these
zero-shot translations by using them as few-shot prompts to sample a
much larger synthetic dataset from a smaller model.
We then \emph{distill} this dataset by discarding the few-shot prompts
and fine-tuning on formatted bitext, producing a language model
aligned with our task format and amenable to backtranslation.
In detail,
we implement the bootstrap as follows:
\begin{enumerate}
\item Generatively pre-train a language model \(p_{\vtheta}(\cdot)\) on a large corpus of
  Internet data.
\item Sample a pool of \(N_S\) synthetic target-side translations and
  \(N_S\) target-side translations zero-shot from another language
  model \(q(\cdot)\) for few-shot prompting. Using \(k\) few-shot
  examples randomly drawn from \(N_S\) (resp. \(N_T\)), sample \(C_S\) synthetic target-side
  translations (resp. \(C_T\) synthetic source-side translations) from
  \(p_{\vtheta}(\cdot)\), using the
  monolingual source-side corpus
  \(\mathcal{M}_S\) (resp. target-side corpus \(\mathcal{M}_T\)).
\item Discard the few-shot prompts, reformat the (gold prompt, sampled
  translation) data as specified in
  \Cref{sec:bts}, and fine-tune the language model
  \(p_{\vtheta}(\cdot)\) on these data.
\item Reverse all data and continue fine-tuning the language model \(p_{\vtheta}(\cdot)\) on the backtranslations (sampled
  translation, gold prompt).
\end{enumerate}

\paragraph{Why amplify and distill?} While few-shot prompting is flexible and
enables strong performance on a diverse suite of NLP tasks
to be coaxed out of generatively pre-trained LMs, its benefits are
most pronounced with larger models, with commensurate training,
inference, compute, and data costs. It is also unclear how to
iteratively fine-tune a language model in a way that preserves its
few-shot ability while remaining aligned with a zero-shot format like in
\Cref{sec:bts}. Few-shot amplification allows us to generate data for the
bootstrap in an unsupervised fashion, possibly avoiding the overhead
of few-shot sampling from GPT-3 itself by few-shot prompting a smaller
model \(p_{\vtheta}(\cdot)\), while distillation enables
iterative backtranslation.

\section{Results}
\label{sec:results}

\paragraph{Experimental setup}

For our experiments, we focus on the well-studied WMT14 English-French
benchmark. In the notation of \Cref{alg:backtranslation}, we obtain
source and target monotext \(\mathcal{M}_S\) and \(\mathcal{M}_T\) by
splitting the WMT14 English-French training set in half, each with
approximately twenty million examples,
and use only the English text from one half and only French text from
the other to avoid implicit sentence-level alignment between source
and target monotext. At each iteration of backtranslation, we sample
one million translations in either direction, \emph{i.e},. \(J=2e6\), and
train for one epoch on the newly sampled data.
For all of our results, unless otherwise specified, we run 40
iterations of backtranslation after the bootstrap and report BLEU
using the final model checkpoint.

To implement the bootstrap, we additionally set aside \(2048\)
training examples, and sample \(N_S=1024\) English-French (resp. \(N_T=1024\)
French-English) translations
zero-shot from GPT-3 to use as few-shot prompts. During few-shot
amplification, we sample four million initial target- and
source-side translations respectively using few-shot prompts,
\emph{i.e.}, \(C_S = C_T = 4e6\) in the notation of \Cref{para:bootstrap},
drawing monolingual prompts from as
\(\mathcal{M}_S\) and \(\mathcal{M}_T\) defined above. We fine-tune for
two epochs in the forwards direction (distillation) and for another
two epochs in the backwards direction (initial backtranslation). For few-shot
prompting, we use \(k=3\) in-context examples. In
\Cref{subsec:few-shot-ablation} we will see that we can minimize the
number of few-shot examples to
\(N_S=N_T=3\) with little effect on evaluation BLEU score after
iterative backtranslation.

We use the same training setup and BPE tokenizer as GPT-3. During fine-tuning, we use a
constant learning rate of \(0.05 \cdot \ell\), where \(\ell\) is the
pre-training learning rate, a weight decay of \(0.1\), and residual
dropout \(0.1\). When sampling during the bootstrap or during
backtranslation, we default to using temperature \(\tau=0.3\). We
ablate other values of \(\tau\) in \Cref{subsec:temp-ablation}. We also
filter all fine-tuning bitext by length, discarding pairs with a
source/target length ratio exceeding \(2.0\).

We report BLEU score on the official WMT14 English-French test set with
greedy (argmax) sampling and sacreBLEU\footnote{Signature:
  \texttt{BLEU+case.mixed+numrefs.1+smooth.exp+tok.intl+version.1.2.20}.}
\citep{DBLP:conf/wmt/Post18}. In \Cref{tab:comparison} we give a comparison to previous
work on unsupervised NMT using \texttt{multi-bleu.perl} and the XLM \citep{DBLP:conf/nips/ConneauL19} tokenizer.

\subsection{Few-shot self-distillation and
  backtranslation} \label{subsec:self-distill}

\begin{table*}[!htbp]

\begin{center}
\begin{tabular}{rrrrrr}
\toprule
 & & \texttt{small} & \texttt{medium}  & \texttt{large} & \texttt{xl}\\
    \cmidrule(r){1-6}
    \multirow{2}{*}{few-shot \((\tau=0.0)\)}
  & en-fr & 1.15 & 7.71 & 13.07 & 14.28  \\
  & fr-en  & 5.04 & 16.87 & 20.25 & 23.0 \\

    \cmidrule(r){1-6}
    \multirow{2}{*}{few-shot \((\tau=0.3)\)}
  & en-fr & 1.02 & 7.36 & 11.89 & 13.58  \\
  & fr-en  & 4.46 & 16.13 & 20.7 & 22.07 \\

    \cmidrule(r){1-6}
    \multirow{2}{*}{few-shot \((\tau=1.0)\)}
  & en-fr & 0.25 & 2.12 & 2.68 & 3.38  \\
  & fr-en  & 1.22 & 5.45 & 6.14 & 9.32 \\

    \cmidrule(r){1-6}
    \multirow{2}{*}{distillation}
  & en-fr & 0.61 & 9.51 & 17.68 & 22.19  \\
  & fr-en  & 4.31 & 23.67 & 29.38 & 31.12 \\

    \cmidrule(r){1-6}
    \multirow{2}{*}{initial backtranslation}
  & en-fr & 7.94 & 29.84 & 33.59 & 34.71  \\
  & fr-en  & 1.5 & 23.12 & 28.58 & 30.52 \\

    \cmidrule(r){1-6}
    \multirow{2}{*}{after backtranslation}
  & en-fr & 30.48 & 36.53 & 37.59 & 39.12  \\
  & fr-en  & 27.24 & 32.15 & 34.79 & 35.43 \\

\bottomrule
\end{tabular}
\end{center}
    \caption{English-French (top) and French-English (bottom) test
    BLEU throughout the few-shot self-distillation bootstrap
  across multiple model scales. } \label{tab:unsup-self-distill-results}
\end{table*}

We first report results using \emph{self-distillation},
\emph{i.e.}, where during the bootstrap
(\Cref{sec:bootstrap}) we sample from a single model which is
then trained to imitate and then backtranslate its own few-shot prompted
generations; for these experiments, the few-shot demonstrations
themselves are
generated zero-shot by GPT-3.
This is then followed by the iterative backtranslation
procedure described in \Cref{sec:bts}.
We apply this methodology to the \texttt{small}, \texttt{medium},
\texttt{large}, and \texttt{xl} models from the GPT-3 family \citep{DBLP:conf/nips/BrownMRSKDNSSAA20}, with
125M, 350M, 760M, and 1.3B parameters respectively.
\Cref{tab:unsup-self-distill-results} displays test BLEU throughout
our procedure
for all model sizes. We see that translation out of English benefits
significantly from the backtranslation part of the bootstrap alone. We
also see that our models are much stronger at the translation task
compared to few-shot prompting after only self-distillation. Finally,
all models benefit significantly from iterative backtranslation, with
English-French BLEU always converging to a slightly higher value than
the reverse direction.

\subsection{Distilling self-amplified GPT-3 into smaller models} \label{subsec:200b-distill}

\begin{table*}[!htbp]
\begin{center}
\begin{tabular}{rrrrrr}
\toprule
 & & \texttt{small} & \texttt{medium}  & \texttt{large} &
                                                               \texttt{xl}
  \\
    \cmidrule(r){1-6}
    \multirow{2}{*}{distillation}
  & en-fr & 34.13 & 36.03 & 37.21 & 37.08  \\
  & fr-en  & 32.34 & 34.96 & 36.12 & 36.34 \\

    \cmidrule(r){1-6}
    \multirow{2}{*}{initial backtranslation}
  & en-fr & 34.71 & 36.31 & 38.89 & 39.05  \\
  & fr-en  & 30.95 & 33.73 & 35.16 & 36.51 \\

    \cmidrule(r){1-6}
    \multirow{2}{*}{after backtranslation}
  & en-fr & 35.62 & 37.79 & 38.91 & 39.79  \\
  & fr-en  & 31.28 & 34.08 & 35.57 & 35.97 \\

    \cmidrule(r){1-6}
    \multirow{2}{*}{after backtranslation (+CC100)}
  & en-fr & 39.02 & 41.31 & 41.97 & \textbf{42.08}  \\
  & fr-en  & 33.43 & 35.69 & 36.85 & \textbf{37.09} \\

\bottomrule
\end{tabular}
\end{center}
    \caption{English-French (top) and French-English (bottom) test
    BLEU throughout the bootstrap and after iterative backtranslation,
    this time
    using generations from self-amplified GPT-3 for
    the bootstrap. We observe the best performance by mixing in
    monotext from the English and French components of the CC100
    dataset \citep{DBLP:conf/lrec/WenzekLCCGJG20,
      DBLP:conf/acl/ConneauKGCWGGOZ20} during backtranslation.}  \label{tab:unsup-200b-results}
\end{table*}

Although we do not apply our full methodology to the 175B parameter GPT-3
model due to compute constraints, we
observe that for few-shot
distillation, instead of training a model on few-shot
samples from itself, we can just as well distill on few-shot samples from a
much larger model instead---in this case, the full-size 175B parameter
GPT-3 model (henceforth just ``GPT-3''). That is, we use GPT-3 to
self-amplify its own zero-shot translations to produce an initial
dataset for distillation.

We now proceed to apply the same method as in
\Cref{subsec:self-distill} to all model sizes,
but this time using few-shot samples from GPT-3 for the bootstrap.
We display the evaluation BLEU scores throughout the
bootstrap and after iterative backtranslation in
\Cref{tab:unsup-200b-results}. Interestingly, the higher-quality
samples from GPT-3 appear to saturate the smaller models and they improve
very little. Motivated by the possibility that our models are beginning to
overfit to the WMT14 English-French training data, we attempt another
experiment where 50\% of the monotext for backtranslation is sampled
from the English and French components of the CC100 dataset
\citep{DBLP:conf/acl/ConneauKGCWGGOZ20}. The extra monolingual data
significantly benefits all model scales,
improving English-French BLEU by approximately 3 points compared to
iterative backtranslation on WMT data alone. With this setup, the \texttt{xl} attains a new
unsupervised state-of-art of 42.1 BLEU on the WMT14 English-French benchmark.

\section{Discussion and further ablations}
\label{sec:discussion}

\paragraph{Bias towards English generation} Previous work
\citep{DBLP:conf/nips/BrownMRSKDNSSAA20} has shown that after
generative pre-training on a corpus of English-dominated Internet
text, GPT-3 models are far more capable of translating into English
than translating out of English. This is reflected by the disparity
between English-French and French-English BLEU scores immediately
after few-shot distillation and before backtranslation on the few-shot
prompted data. Interestingly, after only two epochs of backtranslation
on the relatively scarce few-shot prompted data, this gap is
reversed, with all models achieving significantly higher
English-French BLEU than French-English BLEU. The data efficiency of the bootstrap suggests that coming
out of pre-training, the models are merely misaligned rather than deficient
in knowledge about French, and that their latent
knowledge about translation out of English can be surfaced using backtranslation.
Relatedly, high-quality samples in one language in the previous
round of backtranslation lead to higher-quality synthetic bitext
for training the reverse direction in the next. This turns the
asymmetry towards English generation into an advantage during
backtranslation. However, if the initial disparity between
the quality of the translation directions is extreme (as with the self-distilled
\texttt{small}, which after distillation achieves \(<1\) BLEU for English-French
 versus \(\approx\)\(5\) BLEU for French-English), then we
see that the evaluation BLEU scores for either direction are unstable and
oscillates between iterations,
though they eventually converge upwards as backtranslation continues.  

\paragraph{Comparison to previous work} In \Cref{tab:comparison}, we compare the BLEU scores
attained by our best model (an \texttt{xl} distilled on self-amplified
GPT-3 followed by 40 rounds of backtranslation) to prior work in
unsupervised neural machine translation on the WMT14 English-French
benchmark. To ensure comparability to prior work, we report
  tokenized BLEU using
  \texttt{multi-bleu.perl} and the XLM tokenizer. This was used to
  report the few- and zero-shot performance of GPT-3 in
  \citet{DBLP:conf/nips/BrownMRSKDNSSAA20}, which we also include in
  \Cref{tab:comparison} for completeness.
  We emphasize the improvement of our model
  compared to
  zero-shot GPT-3, which was used to initialize the bootstrap.

\begin{table*}[!hp]
\begin{center}
\begin{tabular}{rccccccccc}

\toprule
&  & XLM & MASS
  &CUNMT  & XLM+ & CBD & \texttt{xl}
  & GPT-3 (fs) & GPT-3 (zs) \\
                                                                                                                                                                                                                                                                                                    
    \cmidrule(r){1-10}
      & en-fr & 33.4 & 37.5 & 37.6 & 40.2  & 38.2 & \textbf{41.7} & 32.6 & 25.2
  \\
  & fr-en  & 33.3 & 34.9 & 35.2  & 36.9  & 35.5 & \textbf{38.0} &
                                                                  \underline{39.2}
      & 21.2 \\
\bottomrule
\end{tabular}
\end{center}
\caption{Comparison of our best model---an \texttt{xl} distilled on
  self-amplified GPT-3 followed by 40 rounds of iterative
  backtranslation---to prior work
  \citep{DBLP:conf/nips/ConneauL19,DBLP:conf/icml/SongTQLL19,DBLP:conf/naacl/WangBLZ21,
    DBLP:journals/tacl/KeungSLS20, DBLP:conf/icml/NguyenJN0A21} in unsupervised NMT on the WMT14
  English-French benchmark.     Bold indicates unsupervised state-of-the-art and underline indicates
  few-shot state-of-the-art.
  } \label{tab:comparison}
\end{table*}

\paragraph{Potential data contamination from pre-training}
For high-resource language pairs such as English-French, naturally
occuring demonstrations of translation are virtually guaranteed to
appear in Common Crawl-like datasets; indeed,
\citet{radford2019language} provide examples of English-French
parallel text embedded in the WebText corpus.
Train/test contamination is also a growing area of concern when
training large language models on internet-derived data. The
data contamination study conducted by
\citet{DBLP:conf/nips/BrownMRSKDNSSAA20} found virtually no test set
contamination for the WMT translation datasets they considered,
including the WMT14 English-French dataset.
We emphasize that
throughout our entire procedure, no explicit supervision is given for
the translation task during pre-training, distillation, or
backtranslation.

\subsection{Ablating temperature for few-shot
  distillation} \label{subsec:temp-ablation}
\begin{table*}[!h]

\begin{center}
\begin{tabular}{rrrrrrr}
\toprule
 & & self-distill & backtrans. \(\tau=0.0\)  & backtrans. \(\tau=0.3\) & backtrans. \(\tau=1.0\) \\
  \cmidrule(r){1-6}
  \multirow{2}{*}{\(\tau=0.0\)}
  & en-fr & 20.3 & 34.4 & 34.7 & 27.8  \\
  & fr-en & 29.9 & 29.3 & 29.6 & 24.7  \\
  \cmidrule(r){1-6}
  \multirow{2}{*}{\(\tau=0.3\)}
  & en-fr & 20.6 & 33.9 & 35.1 & 27.6  \\
  & fr-en & 29.2 & 28.9 & 29.9 & 24.4  \\
  \cmidrule(r){1-6}
  \multirow{2}{*}{\(\tau=1.0\)}
  & en-fr & 20.2 & 34.9 & 34.6 & 27.6  \\
  & fr-en & 29.0 & 29.2 & 29.2 & 24.9  \\
\bottomrule
\end{tabular}
\end{center}
    \caption{English-French (top) and French-English (bottom) test
    BLEU using few-shot prompted samples generated with temperatures
    \(\tau=0.0, 0.3, 1.0\) throughout the bootstrap. We see that the
    temperature used for sampling has little effect on evaluation BLEU
  after few-shot distillation, while high-temperature samples are
  harmful during the backtranslation part of the bootstrap.}
\label{tab:temp-ablation-bleu}
\end{table*}

It was shown by \citet{DBLP:conf/emnlp/EdunovOAG18} that
backtranslation is more effective when the translations are slightly
noisy, \emph{i.e.}, sampled with nonzero temperature or via a noised beam
search.
This motivated our use of the temperature \(\tau=0.3\) throughout. We
ablate this choice of temperature
when sampling data for few-shot distillation, and study the effect of
using \(\tau=0.0\) and \(\tau=1.0\) during the bootstrap using a
\texttt{large} model. We display
the results in \Cref{tab:temp-ablation-bleu}. We see that lower
temperatures lead to marginally higher test BLEU scores during
distillation while \(\tau=1.0\) results in lower test
\emph{loss} and no overfitting after two epochs of training. However,
regardless of the temperature of samples used for self-distillation,
the differences in
both test BLEU and test loss almost vanish after the
backtranslation part of the bootstrap when training to backtranslate
low temperature samples (\(\tau = 0.0\) or \(\tau = 0.3\)).

\subsection{Few-shot self-amplification} \label{subsec:self-amp}

We observed that few-shot prompting GPT-3 with its own zero-shot
translations produced better translations than zero-shot prompting
alone. We investigate this further by comparing the BLEU scores of
zero-shot translations (sampled using
the same prompt described in \Cref{sec:bootstrap}) to the BLEU scores
of self-amplified few-shot
prompted translations (\emph{i.e.}, where the few-shot demonstrations are the
zero-shot translations sampled from the same model) for all the model sizes
studied in this paper. Our results are displayed in
\Cref{tab:self-amp}. We see that self-amplification improves
translation quality at all model scales.

\begin{table*}[!h]

\begin{center}
\begin{tabular}{rrrrrrr}
\toprule
 & & \texttt{small} & \texttt{medium}  & \texttt{large} & \texttt{xl} & GPT-3\\
  \cmidrule(r){1-7}
  \multirow{2}{*}{zero-shot}
  & en-fr & 0.57 & 1.23 & 1.90 & 2.84 & 26.19 \\
  & fr-en & 2.00 & 13.92 & 8.14 & 19.60 & 25.49 \\
  \cmidrule(r){1-7}
  \multirow{2}{*}{self-amplified}
  & en-fr & 1.39 & 8.98 & 12.46 & 14.32 & 29.96 \\
  & fr-en & 5.76 & 16.75 & 21.75 & 23.98 & 31.75 \\
\bottomrule
\end{tabular}
\end{center}
\caption{Zero-shot versus few-shot self-amplified test BLEU for all
  model sizes studied in this paper. For zero-shot generation we use
  the same prompt format described in \Cref{sec:bootstrap}. For
  self-amplified generation, we use the model's own zero-shot
  generations as in-context few-shot examples.}
\label{tab:self-amp}
\end{table*}

\subsection{Using real few-shot examples} \label{subsec:unsup}

\begin{table*}[!h]

\begin{center}
\begin{tabular}{rrrrrr}
\toprule
 & & \texttt{small} & \texttt{medium}  & \texttt{large} & \texttt{xl}\\
\cmidrule(r){1-6}
\multirow{2}{*}{few-shot \((\tau=0.0)\)}
  & en-fr & 1.09 & 7.19 & 11.8 & 13.35  \\
  & fr-en  & 3.86 & 14.58 & 20.34 & 23.01 \\

\cmidrule(r){1-6}
\multirow{2}{*}{few-shot \((\tau=0.3)\)}
  & en-fr & 1.09 & 6.83 & 11.38 & 13.08  \\
  & fr-en  & 4.13 & 14.86 & 19.92 & 22.04 \\

\cmidrule(r){1-6}
\multirow{2}{*}{few-shot \((\tau=1.0)\)}
  & en-fr & 0.33 & 1.74 & 2.34 & 2.94  \\
  & fr-en  & 0.94 & 4.18 & 4.64 & 7.25 \\

\cmidrule(r){1-6}
\multirow{2}{*}{distillation}
  & en-fr & 0.39 & 7.63 & 17.27 & 19.81  \\
  & fr-en  & 3.9 & 20.29 & 27.65 & 30.89 \\

\cmidrule(r){1-6}
\multirow{2}{*}{initial backtranslation}
  & en-fr & 7.77 & 24.71 & 29.64 & 33.78  \\
  & fr-en  & 1.7 & 18.9 & 26.61 & 30.93 \\

\cmidrule(r){1-6}
\multirow{2}{*}{after backtranslation}
  & en-fr & 31.23 & 34.42 & 37.86 & 39.39  \\
  & fr-en  & 27.45 & 29.96 & 34.23 & 34.97 \\

\bottomrule
\end{tabular}
\end{center}
    \caption{English-French (top) and French-English (bottom) test
    BLEU throughout the few-shot self-distillation bootstrap
  across multiple model scales, this time using real few-shot
  examples. We see that performance after backtranslation
  is equivalent to that reported in
  \Cref{tab:unsup-self-distill-results}.} \label{tab:self-distill-results}
\end{table*}

\begin{table*}[!h]
\begin{center}
\begin{tabular}{rrrr}
\toprule
 & & \texttt{small} & \texttt{large}
  \\
\cmidrule(r){1-4}
\multirow{2}{*}{distillation}
  & en-fr & 32.95 & 36.0  \\
  & fr-en  & 32.45 & 36.29 \\

\cmidrule(r){1-4}
\multirow{2}{*}{initial backtranslation}
  & en-fr & 36.32 & 38.72  \\
  & fr-en  & 32.43 & 36.61 \\

\cmidrule(r){1-4}
\multirow{2}{*}{after backtranslation}
  & en-fr & 36.38 & 39.36  \\
  & fr-en  & 32.66 & 35.67 \\

\cmidrule(r){1-4}
\multirow{2}{*}{after backtranslation (+CC100)}
  & en-fr & 39.01 & 42.03  \\
  & fr-en  & 34.17 & 36.94 \\

\bottomrule
\end{tabular}
\end{center}
    \caption{English-French (top) and French-English (bottom) test
    BLEU of the \texttt{small} and \texttt{large} models throughout
    the bootstrap and after iterative backtranslation, where for the
    bootstrap we use generations from 175B GPT-3 prompted using real few-shot examples. Similarly to \Cref{tab:unsup-200b-results}, we observe a boost in
    final BLEU score when, after the bootstrap, we additionally sample monolingual text from
  the English and French portions of the CC100 dataset.} \label{tab:200b-results}
  \end{table*}

So far our results have been completely unsupervised, but
few-shot learning is typically studied in the context of
semi-supervised learning \citep{DBLP:journals/csur/WangYKN20}, where
the few-shot demonstrations are real training data. In this
section, we ablate the usage of synthetic few-shot translations in our
methodology and reproduce our experiments from \Cref{sec:results}
using real few-shot demonstrations. We observe virtually no difference
in BLEU score after iterative backtranslation.

We modify the few-shot prompting described in \Cref{sec:results} as
follows. Rather than sampling zero-shot translations for each half of
our held-out pool of \emph{N}=2048 training examples, we sample from
these examples directly during few-shot prompting.

\Cref{tab:self-distill-results} displays test BLEU throughout the
bootstrap and after iterative backtranslation for the same model sizes
studied in \Cref{subsec:self-distill}. We see that our
models converge to the same test BLEU
(c.f. \Cref{subsec:self-distill}). \Cref{tab:200b-results}
displays analogous results
when distilling samples from GPT-3 with the \texttt{small} and
\texttt{large} models, this time few-shot prompted using real
examples. We again see that using real rather than synthetic few-shot
demonstrations to sample the initial bootstrap data from GPT-3 has no effect on final BLEU score after
iterative backtranslation.

\subsubsection{Almost-unsupervised machine translation with three examples only}

\begin{table*}[!h]
\begin{center}
\begin{tabular*}{\textwidth}{lrrrrrrrrrr}
\toprule
  & \emph{N}=3 & \emph{N}=8 & \emph{N}=16 & \emph{N}=32 & \emph{N}=64 & \emph{N}=128 & \emph{N}=256 & \emph{N}=512 & \emph{N}=1024 & \emph{N}=2048\\
\cmidrule(r){1-11}
en-fr & 12.6 & 12.4 & 12.7 & 13.1 & 13.2 & 13.0 & 12.7 & 12.9 & 12.7 & 12.8\\
fr-en & 21.5 & 21.3 & 22.1 & 22.4 & 21.9 & 22.3 & 22.1 & 22.1 & 22.2 & 22.1\\
\bottomrule
\end{tabular*}
\caption{BLEU scores (calculated over 4096 random training examples) for the
  few-shot prompted translations from a \texttt{large} model, as the total
  number of available few-shot examples varies from \(N=3\) to
  \(N=2048\). We see that \(N\) has minimal impact on the BLEU score
  of the sampled translations. Moreover, the difference in BLEU
  between the models bootstrapped using \(N=3\) versus \(N=2048\) disappears after iterative
  backtranslation. } \label{tab:n-ablation}
\end{center}
\end{table*} 
\label{subsec:few-shot-ablation}
Finally, we show that even in the semi-supervised setting, we can
minimize the supervision available from few-shot demonstrations with
no difference in test BLEU after backtranslation coverges.
\Cref{tab:n-ablation} displays the BLEU scores of few-shot sampled translations
across various orders of magnitude of N, the number of available
few-shot examples. Remarkably, even when N is decreased to 3, there is
only a slight negative impact on the BLEU score of the few-shot sampled
translations. We do not ablate lower values of \(N\) in order to
maintain the assumption of \emph{k}=3 distinct
in-context examples for few-shot prompting. We
then run our entire procedure with a \texttt{large} model,
using \emph{N}=3 real few-shot demonstrations for the bootstrap
followed by iterative backtranslation. We
observe a final English-French BLEU of \(38.0\) and French-English
BLEU of \(34.2\), on par with the final BLEU scores reported in
\Cref{tab:self-distill-results}.

\section{Conclusion and future directions} \label{sec:conclusion}
We remark that backtranslation, like reinforcement learning, is simply a way of
exchanging compute for data. Instead of grounding the model with a
reward signal from an environment, however, backtranslation exploits the symmetry
of the translation task to ground the model by
training it to cross-lingually denoise its own samples.
Our present work can be viewed as part of a
recent trend towards \emph{data-driven architecture engineering}
\citep{DBLP:conf/iclr/Rabe0BS21, DBLP:conf/icml/WuRLBGS21,
  DBLP:journals/corr/abs-2102-07492}, where task-specific
inductive biases, if any, are engineered
into and learned from the training data instead of being hardcoded
into the model architecture. In formulating the translation task in
terms of language modeling, we
see that the input-output inductive bias imposed
by an encoder-decoder architecture can be simulated with prompt
formatting. Similarly, we see that generative language modeling at
sufficient scale combined with clever prompting for automated data generation can attain state-of-the-art results in
unsupervised translation, rendering methods intended to produce
strong encoders and aligned multilingual representations unnecessary.

Although
we have focused solely on the domain of machine translation in this
work, our
methodology is applicable to any sequence-to-sequence task whose forwards and
inverse directions are (1) jointly learnable by an autoregressive
decoder-only transformer and (2) amenable to few-shot prompting after
large-scale generative pre-training. Backtranslation is simply reverse self-training \citep{DBLP:conf/wmt/BojarT11} and is
fundamentally untied to the translation domain; we invite the
research community at large to further explore this technique, moving
beyond translation and towards applications reflecting the full generality of the
transformer architecture.

\bibliography{main}
\bibliographystyle{iclr2022_conference}

\end{document}